\documentclass{article}

\usepackage{amsmath}
\usepackage{amssymb}
\usepackage{amsthm}
\usepackage{placeins}
\usepackage{wrapfig}
\usepackage{subfigure}
\usepackage{nicefrac}

\usepackage{hyperref}

\newtheorem{theorem}{Theorem}
\newtheorem{lemma}{Lemma}

\newcommand{\eg}{e.g.\ }

\newcommand{\E}[2]{\operatorname{\mathbb{E}}_{#1}\left[#2\right]}

\newcommand{\density}{p}

\newcommand{\kl}[2]{\mathrm{D_{KL}}\left(#1\;\middle\|\;#2\right)}
\newcommand{\entropy}{\mathcal{H}}
\newcommand{\ent}{\mathcal{H}}

\newcommand{\voidarg}{{\,\cdot\,}}

\newcommand{\sspace}{\mathcal{S}}
\newcommand{\aspace}{\mathcal{A}}

\newcommand{\state}{\mathbf{s}}

\newcommand{\st}{{\state_t}}

\newcommand{\stp}{{\state_{t+1}}}

\newcommand{\pdyn}{\density}

\newcommand{\action}{\mathbf{a}}

\newcommand{\at}{{\action_t}}

\newcommand{\atp}{{\action_{t+1}}}

\newcommand{\opt}{^*}

\newcommand{\urv}{\mathbf{u}}

\newcommand{\reward}{r}

\newcommand{\rmin}{r_\mathrm{min}}
\newcommand{\rmax}{r_\mathrm{max}}

\newcommand{\V}{V}

\newcommand{\Q}{Q}

\newcommand{\policy}{\pi}
\newcommand{\policyold}{{\policy_\mathrm{old}}}
\newcommand{\policynew}{{\policy_\mathrm{new}}}

\newcommand{\policyopt}{{\policy^*}}

\newcommand{\params}{\theta}

\newcommand{\pparams}{{\phi}}   
\newcommand{\vparams}{{\psi}}   
\newcommand{\vtargetparams}{{\bar\psi}}   

\newcommand{\gauss}{\mathcal{N}}

\newcommand{\inv}{^{-1}}

\newcommand{\reals}{\mathbb{R}}

\newcommand{\discount}{\gamma}

\newcommand{\aref}[1]{\hyperref[#1]{Appendix~\ref*{#1}}}

\usepackage{microtype}
\usepackage{graphicx}
\usepackage{subfigure}
\usepackage{booktabs} %

\usepackage[accepted]{icml2018}

\icmltitlerunning{Soft Actor-Critic}

\begin{document}

\twocolumn[
\icmltitle{{\LARGE Soft Actor-Critic:} \\ Off-Policy Maximum Entropy Deep Reinforcement\\ 
Learning with a Stochastic Actor}

\icmlsetsymbol{equal}{*}

\begin{icmlauthorlist}
\icmlauthor{Tuomas Haarnoja}{berkeley}
\icmlauthor{Aurick Zhou}{berkeley}
\icmlauthor{Pieter Abbeel}{berkeley}
\icmlauthor{Sergey Levine}{berkeley}
\end{icmlauthorlist}

\icmlaffiliation{berkeley}{Berkeley Artificial Intelligence Research, University of California, Berkeley, USA}
\icmlcorrespondingauthor{Tuomas Haarnoja}{haarnoja@berkeley.edu}

\icmlkeywords{reinforcement learning, control as inference}

\vskip 0.3in
]

\printAffiliationsAndNotice{}  %

\begin{abstract}
Model-free deep reinforcement learning (RL) algorithms have been demonstrated on a range of challenging decision making and control tasks. However, these methods typically suffer from two major challenges: very high sample complexity and brittle convergence properties, which necessitate meticulous hyperparameter tuning. Both of these challenges severely limit the applicability of such methods to complex, real-world domains. In this paper, we propose soft actor-critic, an off-policy actor-critic deep RL algorithm based on the maximum entropy reinforcement learning framework. In this framework, the actor aims to maximize expected reward while also maximizing entropy. That is, to succeed at the task while acting as randomly as possible. Prior deep RL methods based on this framework have been formulated as Q-learning methods. By combining off-policy updates with a stable stochastic actor-critic formulation, our method achieves state-of-the-art performance on a range of continuous control benchmark tasks, outperforming prior on-policy and off-policy methods. Furthermore, we demonstrate that, in contrast to other off-policy algorithms, our approach is very stable, achieving very similar performance across different random seeds.
\end{abstract}

\section{Introduction}

Model-free deep reinforcement learning (RL) algorithms have been applied in a range of challenging domains, from games~\citep{mnih2013playing,silver2016mastering} to robotic control~\citep{schulman2015trust}. The combination of RL and high-capacity function approximators such as neural networks holds the promise of automating a wide range of decision making and control tasks, but widespread adoption of these methods in real-world domains has been hampered by two major challenges. First, model-free deep RL methods are notoriously expensive in terms of their sample complexity. Even relatively simple tasks can require millions of steps of data collection, and complex behaviors with high-dimensional observations might need substantially more. Second, these methods are often brittle with respect to their hyperparameters: learning rates, exploration constants, and other settings must be set carefully for different problem settings to achieve good results. Both of these challenges severely limit the applicability of model-free deep RL to real-world tasks.

One cause for the poor sample efficiency of deep RL methods is on-policy learning: some of the most commonly used deep RL algorithms, such as TRPO~\cite{schulman2015trust}, PPO~\citep{schulman2017proximal} or A3C~\citep{mnih2016asynchronous}, require new samples to be collected for each gradient step. This quickly becomes extravagantly expensive, as the number of gradient steps and samples per step needed to learn an effective policy increases with task complexity. Off-policy algorithms aim to reuse past experience. This is not directly feasible with conventional policy gradient formulations, but is relatively straightforward for Q-learning based methods~\citep{mnih2015human}. Unfortunately, the combination of off-policy learning and high-dimensional, nonlinear function approximation with neural networks presents a major challenge for stability and convergence~\citep{bhatnagar2009convergent}.
This challenge is further exacerbated in continuous state and action spaces, where a separate actor network is often used to perform the maximization in Q-learning. A commonly used algorithm in such settings, deep deterministic policy gradient (DDPG)~\citep{lillicrap2015continuous}, provides for sample-efficient learning but is notoriously challenging to use due to its extreme brittleness and hyperparameter sensitivity~\citep{duan2016benchmarking,henderson2017deep}.

We explore how to design an efficient and stable model-free deep RL algorithm for continuous state and action spaces. To that end, we draw on the maximum entropy framework, which augments the standard maximum reward reinforcement learning objective with an entropy maximization term~\citep{ziebart2008maximum,toussaint2009robot,rawlik2012stochastic,fox2015taming,haarnoja2017reinforcement}.
Maximum entropy reinforcement learning alters the RL objective, though the original objective can be recovered using a temperature parameter~\citep{haarnoja2017reinforcement}.
More importantly, the maximum entropy formulation provides a substantial improvement in exploration and robustness: as discussed by \citet{ziebart2010modeling}, maximum entropy policies are robust in the face of model and estimation errors, and as demonstrated by~\cite{haarnoja2017reinforcement}, they improve exploration by acquiring diverse behaviors. Prior work has proposed model-free deep RL algorithms that perform on-policy learning with entropy maximization~\citep{o2016pgq}, as well as off-policy methods based on soft Q-learning and its variants~\citep{schulman2017equivalence, nachum2017bridging, haarnoja2017reinforcement}. However, the on-policy variants suffer from poor sample complexity for the reasons discussed above, while the off-policy variants require complex approximate inference procedures in continuous action spaces.

In this paper, we demonstrate that we can devise an off-policy maximum entropy actor-critic algorithm, which we call soft actor-critic (SAC), which provides for both sample-efficient learning and stability. This algorithm extends readily to very complex, high-dimensional tasks, such as the Humanoid benchmark~\citep{duan2016benchmarking} with 21 action dimensions, where off-policy methods such as DDPG typically struggle to obtain good results~\citep{gu2016q}. SAC also avoids the complexity and potential instability associated with approximate inference in prior off-policy maximum entropy algorithms based on soft Q-learning~\citep{haarnoja2017reinforcement}. We present a convergence proof for policy iteration in the maximum entropy framework, and then introduce a new algorithm based on an approximation to this procedure that can be practically implemented with deep neural networks, which we call soft actor-critic. We present empirical results that show that soft actor-critic attains a substantial improvement in both performance and sample efficiency over both off-policy and on-policy prior methods. We also compare to twin delayed deep deterministic (TD3) policy gradient algorithm~\citep{fujimoto2018addressing}, which is a concurrent work that proposes a deterministic algorithm that substantially improves on DDPG.

\section{Related Work}

Our soft actor-critic algorithm incorporates three key ingredients: an actor-critic architecture with separate policy and value function networks, an off-policy formulation that enables reuse of previously collected data for efficiency, and entropy maximization to enable stability and exploration. We review prior works that draw on some of these ideas in this section. Actor-critic algorithms are typically derived starting from policy iteration, which alternates between \emph{policy evaluation}---computing the value function for a policy---and \emph{policy improvement}---using the value function to obtain a better policy~\citep{barto1983neuronlike,sutton1998reinforcement}. In large-scale reinforcement learning problems, it is typically impractical to run either of these steps to convergence, and instead the value function and policy are optimized jointly. In this case, the policy is referred to as the actor, and the value function as the critic. Many actor-critic algorithms build on the standard, on-policy policy gradient formulation to update the actor~\citep{peters2008reinforcement}, and many of them also consider the entropy of the policy, but instead of maximizing the entropy, they use it as an regularizer \citep{schulman2017proximal,schulman2015trust,mnih2016asynchronous,gruslys2017reactor}. On-policy training tends to improve stability but results in poor sample complexity.

There have been efforts to increase the sample efficiency while retaining robustness by incorporating off-policy samples and by using higher order variance reduction techniques~\citep{o2016pgq,gu2016q}. However, fully off-policy algorithms still attain better efficiency. A particularly popular off-policy actor-critic method, DDPG~\cite{lillicrap2015continuous}, which is a deep variant of the deterministic policy gradient~\citep{silver2014deterministic} algorithm, uses a Q-function estimator to enable off-policy learning, and a deterministic actor that maximizes this Q-function. As such, this method can be viewed both as a deterministic actor-critic algorithm and an approximate Q-learning algorithm. Unfortunately, the interplay between the deterministic actor network and the Q-function typically makes DDPG extremely difficult to stabilize and brittle to hyperparameter settings~\citep{duan2016benchmarking,henderson2017deep}. As a consequence, it is difficult to extend DDPG to complex, high-dimensional tasks, and on-policy policy gradient methods still tend to produce the best results in such settings~\citep{gu2016q}. Our method instead combines off-policy actor-critic training with a \emph{stochastic} actor, and further aims to maximize the entropy of this actor with an entropy maximization objective. We find that this actually results in a considerably more stable and scalable algorithm that, in practice, exceeds both the efficiency and final performance of DDPG. A similar method can be derived as a zero-step special case of stochastic value gradients (SVG(0))~\cite{heess2015learning}. However, SVG(0) differs from our method in that it optimizes the standard maximum expected return objective, and it does not make use of a separate value network, which we found to make training more stable.

Maximum entropy reinforcement learning optimizes policies to maximize both the expected return and the expected entropy of the policy. This framework has been used in many contexts, from inverse reinforcement learning~\citep{ziebart2008maximum} to optimal control~\citep{todorov2008general,toussaint2009robot,rawlik2012stochastic}. In guided policy search \citep{levine2013guided,levine2016end}, the maximum entropy distribution is used to guide policy learning towards high-reward regions. More recently, several papers have noted the connection between Q-learning and policy gradient methods in the framework of maximum entropy learning~\citep{o2016pgq,haarnoja2017reinforcement,nachum2017bridging,schulman2017equivalence}. While most of the prior model-free works assume a discrete action space, \citet{nachum2017trust} approximate the maximum entropy distribution with a Gaussian and \citet{haarnoja2017reinforcement} with a sampling network trained to draw samples from the optimal policy. Although the soft Q-learning algorithm proposed by \citet{haarnoja2017reinforcement} has a value function and actor network, it is not a true actor-critic algorithm: the Q-function is estimating the optimal Q-function, and the actor does not directly affect the Q-function except through the data distribution. Hence, \citet{haarnoja2017reinforcement} motivates the actor network as an approximate sampler, rather than the actor in an actor-critic algorithm. Crucially, the convergence of this method hinges on how well this sampler approximates the true posterior. In contrast, we prove that our method converges to the optimal policy from a given policy class, regardless of the policy parameterization. Furthermore, these prior maximum entropy methods generally do not exceed the performance of state-of-the-art off-policy algorithms, such as DDPG, when learning from scratch, though they may have other benefits, such as improved exploration and ease of fine-tuning. In our experiments, we demonstrate that our soft actor-critic algorithm does in fact exceed the performance of prior state-of-the-art off-policy deep RL methods by a wide margin.

\vspace{-1mm}
\section{Preliminaries}
\label{sec:preliminaries}

We first introduce notation and summarize the standard and maximum entropy reinforcement learning frameworks.

\vspace{-1mm}
\subsection{Notation}

We address policy learning in continuous action spaces. We consider an infinite-horizon Markov decision process (MDP), defined by the tuple $(\sspace, \aspace, \pdyn, \reward)$, where the state space $\sspace$ and the action space $\aspace$ are continuous, and the unknown state transition probability $\pdyn:\ \sspace \times \sspace \times \aspace \rightarrow [0,\, \infty)$ represents the probability density of the next state $\stp\in\sspace$ given the current state $\st\in\sspace$ and action $\at\in\aspace$. The environment emits a bounded reward $\reward: \sspace \times \aspace \rightarrow  [\rmin,\rmax]$ on each transition. We will use $\rho_\policy(\st)$ and $\rho_\policy(\st,\at)$ to denote the state and state-action marginals of the trajectory distribution induced by a policy $\policy(\at|\st)$. 

\subsection{Maximum Entropy Reinforcement Learning}

Standard RL maximizes the expected sum of rewards $\sum_t \E{(\st,\at)\sim\rho_\policy}{\reward(\st,\at)}$. We will consider a more general maximum entropy objective (see \eg \citet{ziebart2010modeling}), which favors stochastic policies by augmenting the objective with the expected entropy of the policy over $\rho_\policy(\st)$:
\begin{align}
\label{eq:maxent_objective}
J(\policy)  = \sum_{t=0}^{T} \E{(\st, \at) \sim \rho_\policy}{\reward(\st,\at) + \alpha\ent(\policy(\voidarg|\st))}.
\end{align}
The temperature parameter $\alpha$ determines the relative importance of the entropy term against the reward, and thus controls the stochasticity of the optimal policy. The maximum entropy objective differs from the standard maximum expected reward objective used in conventional reinforcement learning, though the conventional objective can be recovered in the limit as $\alpha \rightarrow 0$. For the rest of this paper, we will omit writing the temperature explicitly, as it can always be subsumed into the reward by scaling it by $\alpha\inv$. 

This objective has a number of conceptual and practical advantages. First, the policy is incentivized to explore more widely, while giving up on clearly unpromising avenues. Second, the policy can capture multiple modes of near-optimal behavior. In problem settings where multiple actions seem equally attractive, the policy will commit equal probability mass to those actions.
Lastly, prior work has observed improved exploration with this objective~\citep{haarnoja2017reinforcement,schulman2017equivalence}, and in our experiments, we observe that it considerably improves learning speed over state-of-art methods that optimize the conventional RL objective function. We can extend the objective to infinite horizon problems by introducing a discount factor $\discount$ to ensure that the sum of expected rewards and entropies is finite. Writing down the maximum entropy objective for the infinite horizon discounted case is more involved~\citep{thomas2014bias} and is deferred to \aref{app:objective}.

Prior methods have proposed directly solving for the optimal Q-function, from which the optimal policy can be recovered~\citep{ziebart2008maximum,fox2015taming,haarnoja2017reinforcement}. We will discuss how we can devise a soft actor-critic algorithm through a policy iteration formulation, where we instead evaluate the Q-function of the current policy and update the policy through an \emph{off-policy} gradient update. Though such algorithms have previously been proposed for conventional reinforcement learning, our method is, to our knowledge, the first off-policy actor-critic method in the maximum entropy reinforcement learning framework.

\section{From Soft Policy Iteration to Soft Actor-Critic} 
\label{sec:soft_policy_iteration}

Our off-policy soft actor-critic algorithm can be derived starting from a maximum entropy variant of the policy iteration method. We will first present this derivation, verify that the corresponding algorithm converges to the optimal policy from its density class, and then present a practical deep reinforcement learning algorithm based on this theory.

\subsection{Derivation of Soft Policy Iteration}

We will begin by deriving soft policy iteration, a general algorithm for learning optimal maximum entropy policies that alternates between policy evaluation and policy improvement in the maximum entropy framework. Our derivation is based on a tabular setting, to enable theoretical analysis and convergence guarantees, and we extend this method into the general continuous setting in the next section. We will show that soft policy iteration converges to the optimal policy within a set of policies which might correspond, for instance, to a set of parameterized densities.

In the policy evaluation step of soft policy iteration, we wish to compute the value of a policy $\policy$ according to the maximum entropy objective in~\autoref{eq:maxent_objective}. For a fixed policy, the soft Q-value can be computed iteratively, starting from any function $Q: \sspace\times \aspace \rightarrow \reals$ and repeatedly applying a modified Bellman backup operator $\mathcal{T}^\policy$ given by
\begin{align}
\label{eq:soft_bellman_backup_op}
\mathcal{T}^\policy Q(\st, \at) \triangleq  \reward(\st, \at) + \discount \E{\stp \sim \pdyn}{V(\stp)},
\end{align}
where
\begin{align}
V(\st) = \E{\at\sim\policy}{\Q(\st, \at) - \log\policy(\at|\st)}
\label{eq:soft_value_function}
\end{align}
is the soft state value function. We can obtain the soft value function for any policy $\policy$ by repeatedly applying $\mathcal{T}^\policy$ as formalized below.
\begin{lemma}[Soft Policy Evaluation]
\label{lem:soft_policy_evaluation}
Consider the soft Bellman backup operator $\mathcal{T}^\policy$ in \autoref{eq:soft_bellman_backup_op} and a mapping $Q^0: \sspace \times \aspace\rightarrow \reals$ with $|\aspace|<\infty$, and define $\Q^{k+1} = \mathcal{T}^\policy \Q^k$. Then the sequence $Q^k$ will converge to the soft Q-value of $\policy$ as $k\rightarrow \infty$.
\begin{proof}
See \aref{app:lem_soft_policy_evaluation}.
\end{proof}
\end{lemma}
In the policy improvement step, we update the policy towards the exponential of the new Q-function. This particular choice of update can be guaranteed to result in an improved policy in terms of its soft value.
Since in practice we prefer policies that are tractable, we will additionally restrict the policy to some set of policies $\Pi$, which can correspond, for example, to a parameterized family of distributions such as Gaussians. 
To account for the constraint that $\policy \in \Pi$, we project the improved policy into the desired set of policies. While in principle we could choose any projection, it will turn out to be convenient to use the information projection defined in terms of the Kullback-Leibler divergence. In the other words, in the policy improvement step, for each state, we update the policy according to
\begin{align}
\policy_\mathrm{new} = \arg\underset{\policy'\in \Pi}{\min}\kl{\policy'(\voidarg|\st)}{\frac{\exp\left(Q^{\policy_\mathrm{old}}(\st, \voidarg)\right)}{Z^{\policy_\mathrm{old}}(\st)}}.
\label{eq:constrainted_policy_fitting}
\end{align}
The partition function $Z^{\policy_\mathrm{old}}(\st)$ normalizes the distribution, and while it is intractable in general, it does not contribute to the gradient with respect to the new policy and can thus be ignored, as noted in the next section. For this projection, we can show that the new, projected policy has a higher value than the old policy with respect to the objective in~\autoref{eq:maxent_objective}. We formalize this result in \autoref{lem:policy_improvement}.
\begin{lemma}[Soft Policy Improvement]
\label{lem:policy_improvement}
Let $\policy_\mathrm{old} \in \Pi$ and let $\policy_\mathrm{new}$ be the optimizer of the minimization problem defined in \autoref{eq:constrainted_policy_fitting}. Then $\Q^{\policy_\mathrm{new}}(\st, \at) \geq \Q^{\policy_\mathrm{old}}(\st, \at)$ for all $(\st, \at) \in \sspace\times\aspace$ with $|\aspace|<\infty$.
\begin{proof}
See \aref{app:lem_policy_improvement}.
\end{proof}
\end{lemma}
The full soft policy iteration algorithm alternates between the soft policy evaluation and the soft policy improvement steps, and it will provably converge to the optimal maximum entropy policy among the policies in $\Pi$ (\autoref{the:soft_policy_iteration}). Although this algorithm will provably find the optimal solution, we can perform it in its exact form only in the tabular case. Therefore, we will next approximate the algorithm for continuous domains, where we need to rely on a function approximator to represent the Q-values, and running the two steps until convergence would be computationally too expensive. The approximation gives rise to a new practical algorithm, called soft actor-critic.
\begin{theorem}[Soft Policy Iteration]
\label{the:soft_policy_iteration}
Repeated application of soft policy evaluation and soft policy improvement from any $\policy\in\Pi$ converges to a policy $\policy\opt$ such that $Q^{\policy\opt}(\st, \at) \geq Q^{\policy}(\st, \at)$ for all $\policy \in \Pi$ and $(\st, \at) \in \sspace\times\aspace$, assuming $|\aspace|<\infty$.
\begin{proof}
See \aref{app:the_soft_policy_iteration}.
\end{proof}
\end{theorem}

\subsection{Soft Actor-Critic}

As discussed above, large continuous domains require us to derive a practical approximation to soft policy iteration. To that end, we will use function approximators for both the Q-function and the policy, and instead of running evaluation and improvement to convergence, alternate between optimizing both networks with stochastic gradient descent.
We will consider a parameterized state value function $\V_\vparams(\st)$, soft Q-function $\Q_\params(\st, \at)$, and a tractable policy $\policy_\pparams(\at|\st)$. The parameters of these networks are $\vparams,\ \params$, and $\pparams$.
For example, the value functions can be modeled as expressive neural networks, and the policy as a Gaussian with mean and covariance given by neural networks. We will next derive update rules for these parameter vectors.

The state value function approximates the soft value. There is no need in principle to include a separate function approximator for the state value, since it is related to the Q-function and policy according to \autoref{eq:soft_value_function}. This quantity can be estimated from a single action sample from the current policy without introducing a bias, but in practice, including a separate function approximator for the soft value can stabilize training
and is convenient to train simultaneously with the other networks. The soft value function is trained to minimize the squared residual error 
\begin{align}
\label{eq:v_cost}
\resizebox{\columnwidth}{!}{$
J_V(\vparams) = \E{\st \sim \mathcal{D}}{\frac{1}{2}\left(\V_\vparams(\st) - \E{\at\sim\policy_\pparams}{Q_\params(\st, \at) - \log \policy_\pparams(\at|\st)}\right)^2}\,$}
\end{align}
where $\mathcal{D}$ is the distribution of previously sampled states and actions, or a replay buffer. The gradient of \autoref{eq:v_cost} can be estimated with an unbiased estimator
\begin{align}
\resizebox{\columnwidth}{!}{$
\hat \nabla_\vparams J_V(\vparams) = \nabla_\vparams \V_\vparams(\st) \left(\V_\vparams(\st) - Q_\params(\st, \at) + \log \policy_\pparams(\at|\st)\right),$}
\label{eq:v_gradient}
\end{align}
where the actions are sampled according to the current policy, instead of the replay buffer. The soft Q-function parameters can be trained to minimize the soft Bellman residual
\begin{align}
J_\Q(\params) = \E{(\st, \at)\sim\mathcal{D}}{\frac{1}{2}\left(\Q_\params(\st, \at) - \hat \Q(\st, \at)\right)^2},
\label{eq:q_cost}
\end{align}
with 
\begin{align}
\hat \Q(\st, \at) = \reward(\st, \at) + \discount \E{\stp\sim\pdyn}{\V_\vtargetparams(\stp)},
\end{align}
which again can be optimized with stochastic gradients
\begin{align}
\resizebox{\columnwidth}{!}{$
\hat \nabla_\params J_Q(\params) =  \nabla_\params \Q_\params(\at, \st) \left(\Q_\params(\st, \at) - \reward(\st, \at) - \discount \V_\vtargetparams(\stp)\right)$}.
\end{align}
The update  makes use of a target value network $V_{\bar\psi}$, where $\bar\psi$ can be an exponentially moving average of the value network weights, which has been shown to stabilize training~\citep{mnih2015human}. Alternatively, we can update the target weights to match the current value function weights periodically (see \autoref{app:benchmarks}).  Finally, the policy parameters can be learned by directly minimizing the expected KL-divergence in \autoref{eq:constrainted_policy_fitting}:
\begin{align}
J_\policy(\pparams) = \E{\st\sim\mathcal{D}}{\kl{\policy_\pparams(\voidarg|\st)}{\frac{\exp\left(Q_\params(\st, \voidarg)\right)}{Z_\params(\st)}}}.
\label{eq:policy_objective}
\end{align}
There are several options for minimizing $J_\policy$. A typical solution for policy gradient methods is to use the likelihood ratio gradient estimator~\citep{williams1992simple}, which does not require backpropagating the gradient through the policy and the target density networks. However, in our case, the target density is the Q-function, which is represented by a neural network an can be differentiated, and it is thus convenient to apply the reparameterization trick instead, resulting in a lower variance estimator. To that end, we reparameterize the policy using a neural network transformation 
\begin{align}
\at = f_\pparams(\epsilon_t; \st),
\end{align}
where $\epsilon_t$ is an input noise vector, sampled from some fixed distribution, such as a spherical Gaussian. We can now rewrite the objective in~\autoref{eq:policy_objective} as
\begin{align}
\resizebox{\columnwidth}{!}{$
J_\policy(\pparams) = \E{\st\sim\mathcal{D},\epsilon_t\sim\gauss}{\log \policy_\pparams(f_\pparams(\epsilon_t;\st)|\st) - Q_\params(\st, f_\pparams(\epsilon_t;\st))},$}
\label{eq:reparam_objective}
\end{align}
where $\policy_\pparams$ is defined implicitly in terms of $f_\pparams$, and we have noted that the partition function is independent of $\pparams$ and can thus be omitted. We can approximate the gradient of~\autoref{eq:reparam_objective} with
\begin{align}
&\hat\nabla_\pparams J_\policy(\pparams) = \nabla_\pparams \log \policy_\pparams(\at|\st) \notag\\
&\ \ \ \ \ \ \ + (\nabla_\at \log \policy_\pparams(\at|\st)
- \nabla_\at Q(\st, \at))\nabla_\pparams f_\pparams(\epsilon_t;\st),
\label{eq:policy_gradient}
\end{align}
where $\at$ is evaluated at $f_\pparams(\epsilon_t; \st)$. This unbiased gradient estimator extends the DDPG style policy gradients~\citep{lillicrap2015continuous} to any tractable stochastic policy.

\begin{algorithm}[tb]
\caption{Soft Actor-Critic}
\label{alg:soft_actor_critic}
\begin{algorithmic}
\STATE \mbox{Initialize parameter vectors $\vparams$, $\vtargetparams$, $\params$, $\pparams$.}
\FOR{each iteration}
	\FOR{each environment step}
	\STATE $\at \sim \policy_\pparams(\at|\st)$
	\STATE $\stp \sim \pdyn(\stp| \st, \at)$
	\STATE $\mathcal{D} \leftarrow \mathcal{D} \cup \left\{(\st, \at, \reward(\st, \at), \stp)\right\}$
	\ENDFOR
	\FOR{each gradient step}
	\item $\vparams \leftarrow \vparams - \lambda_V \hat \nabla_\vparams J_\V(\vparams)$
	\STATE $\params_i \leftarrow \params_i - \lambda_Q \hat \nabla_{\params_i} J_\Q(\params_i)$ for $i\in\{1, 2\}$
	\STATE $\pparams \leftarrow \pparams - \lambda_\policy \hat \nabla_\pparams J_\policy(\pparams)$
	\STATE $\vtargetparams\leftarrow \tau \vparams + (1-\tau)\vtargetparams$
	\ENDFOR
\ENDFOR
\end{algorithmic}
\end{algorithm}

\begin{figure*}[t]
    \centering
    \subfigure[Hopper-v1]{
        \includegraphics[width=0.31\textwidth, trim={0 0 5mm 7.5mm}, clip]{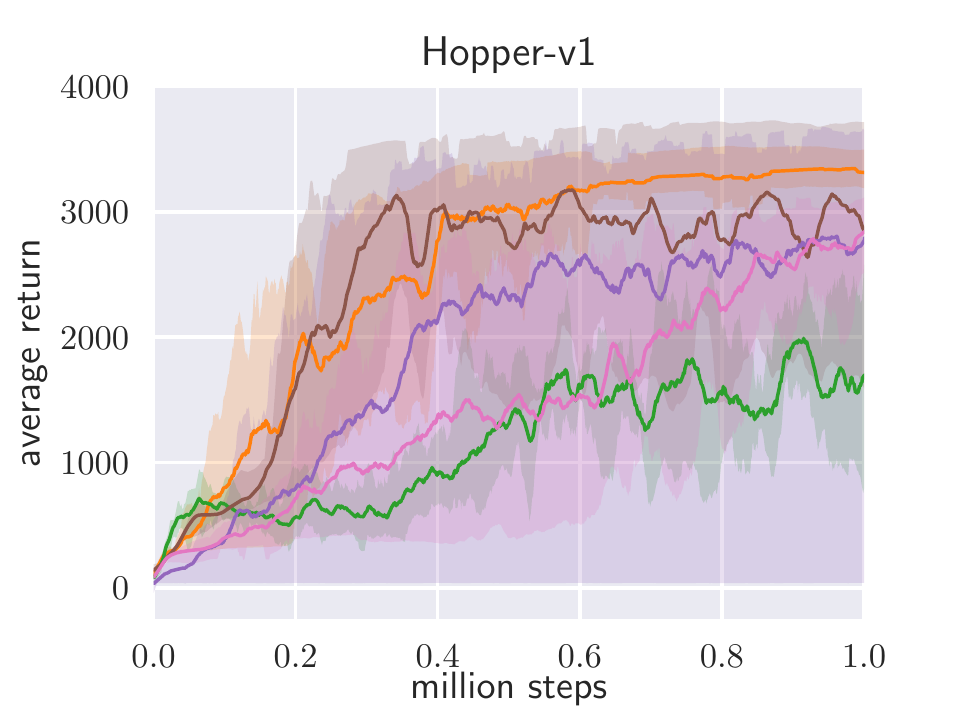}
    }
  	\subfigure[Walker2d-v1]{
        \includegraphics[width=0.31\textwidth, trim={0 0 5mm 7.5mm}, clip]{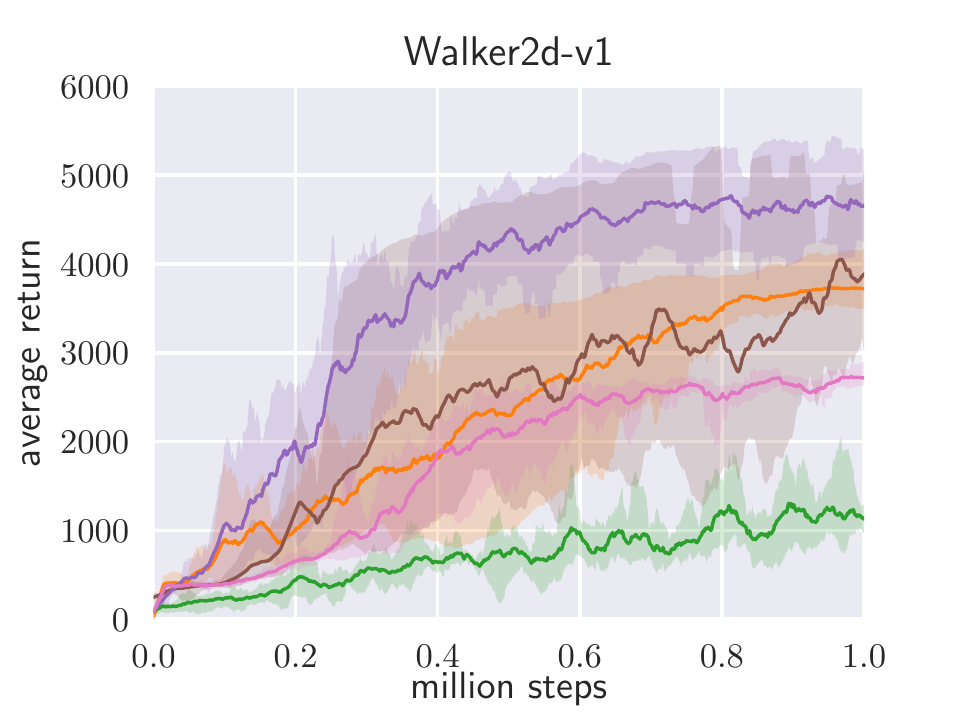}
    }
    \subfigure[HalfCheetah-v1]{
        \includegraphics[width=0.31\textwidth, trim={0 0 5mm 7.5mm}, clip]{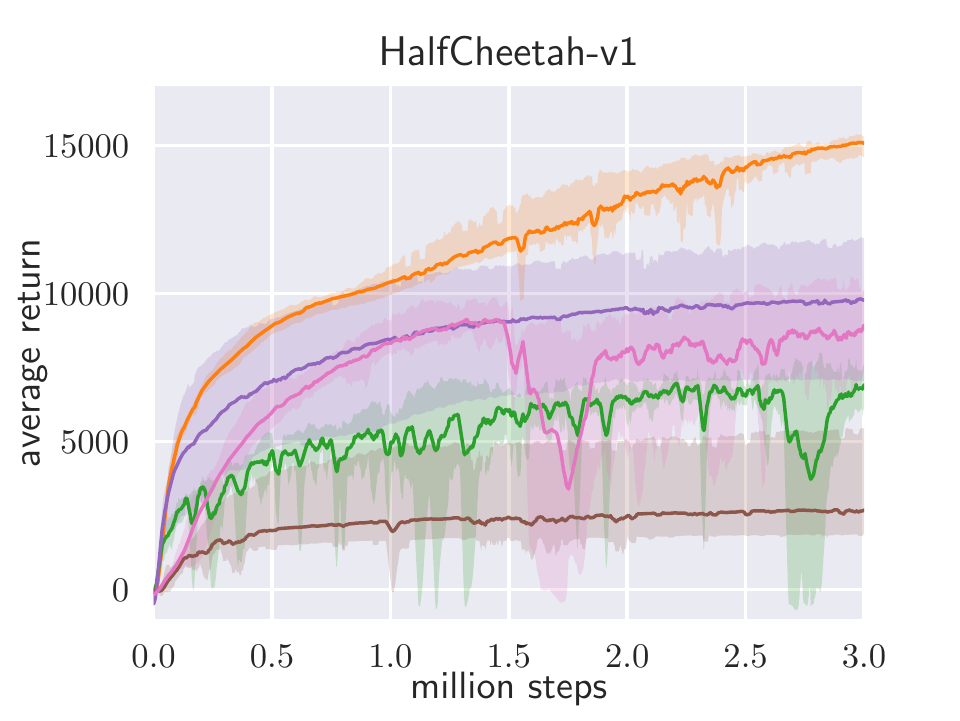}
    }
    \subfigure[Ant-v1]{
        \includegraphics[width=0.31\textwidth, trim={0 0 5mm 7.5mm}, clip]{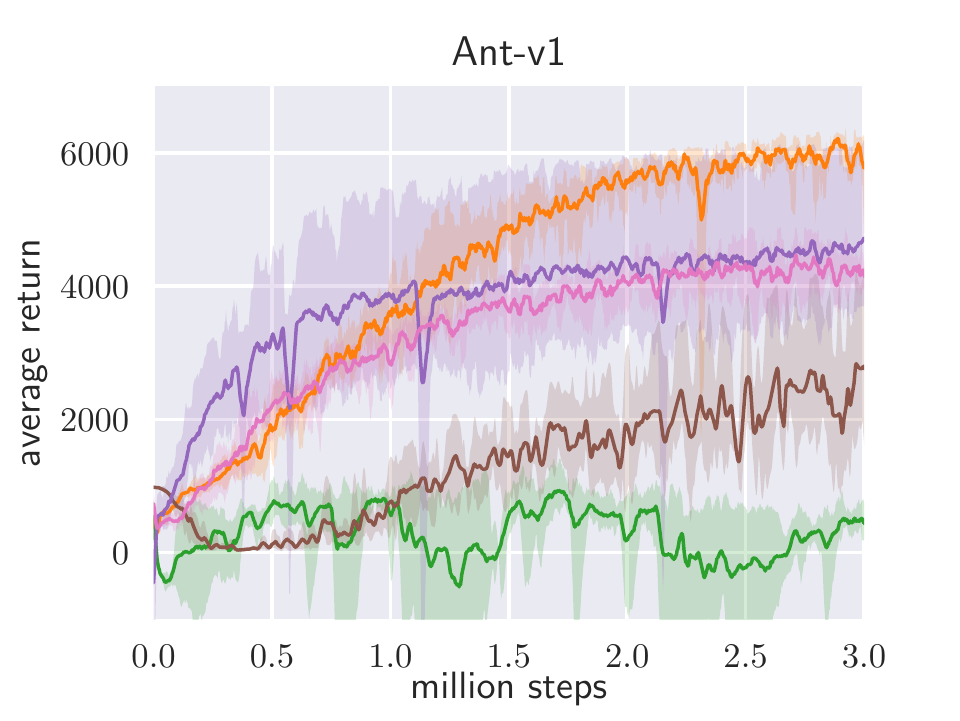}
    }
    \subfigure[Humanoid-v1]{
        \includegraphics[width=0.31\textwidth, trim={0 0 5mm 7.5mm}, clip]{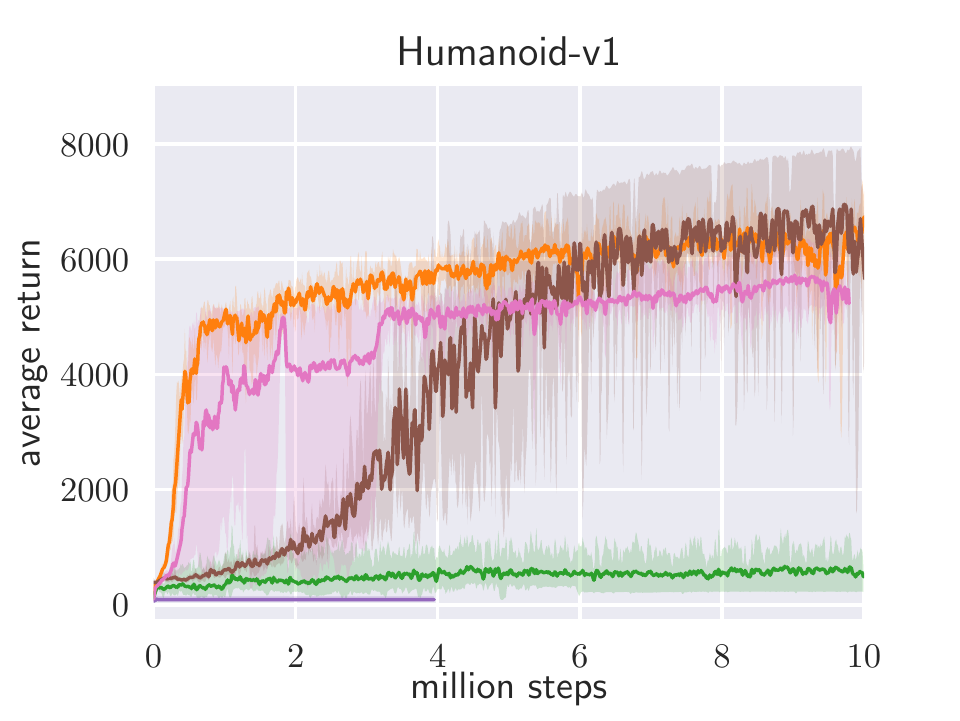}
    }
    \subfigure[Humanoid (rllab)]{
        \includegraphics[width=0.31\textwidth, trim={0 0 5mm 7.5mm}, clip]{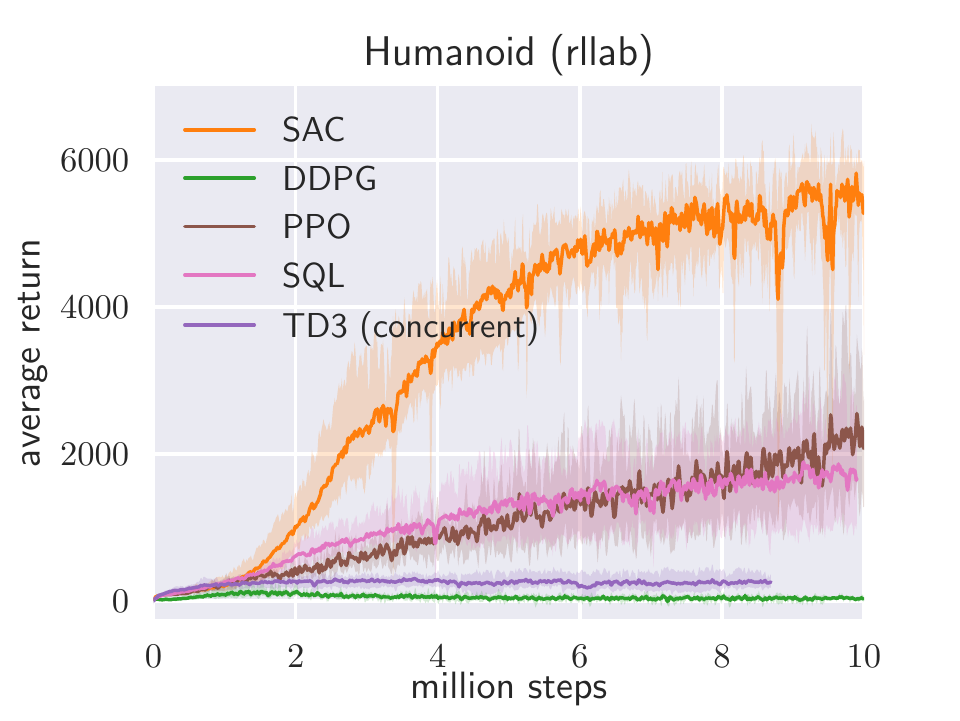}
    }
    \caption{\small Training curves on continuous control benchmarks. Soft actor-critic (yellow) performs consistently across all tasks and outperforming both on-policy and off-policy methods in the most challenging  tasks.}
	\label{fig:training_curves}
	\hspace{2mm}
 \end{figure*}

Our algorithm also makes use of two Q-functions to mitigate  positive bias in the policy improvement step that is known to degrade performance of value based methods~\citep{hasselt2010double,fujimoto2018addressing}. In particular, we parameterize two Q-functions, with parameters $\params_i$, and train them independently to optimize $J_Q(\params_i)$. We then use the minimum of the Q-functions for the value gradient in \autoref{eq:v_gradient} and policy gradient in \autoref{eq:policy_gradient}, as proposed by \citet{fujimoto2018addressing}. Although our algorithm can learn challenging tasks, including a 21-dimensional Humanoid, using just a single Q-function, we found two Q-functions significantly speed up training, especially on harder tasks. The complete algorithm is described in \autoref{alg:soft_actor_critic}. The method alternates between collecting experience from the environment with the current policy and updating the function approximators using the stochastic gradients from batches sampled from a replay buffer. In practice, we take a single environment step followed by one or several gradient steps (see \aref{app:hypers} for all hyperparameter). Using off-policy data from a replay buffer is feasible because both value estimators and the policy can be trained entirely on off-policy data. The algorithm is agnostic to the parameterization of the policy, as long as it can be evaluated for any arbitrary state-action tuple.

\section{Experiments}
\label{sec:experiments}

The goal of our experimental evaluation is to understand how the sample complexity and stability of our method compares with prior off-policy and on-policy deep reinforcement learning algorithms. We compare our method to prior techniques on a range of challenging continuous control tasks from the OpenAI gym benchmark suite~\citep{brockman2016openai} and also on the rllab implementation of the Humanoid task~\citep{duan2016benchmarking}. Although the easier tasks can be solved by a wide range of different algorithms, the more complex benchmarks, such as the 21-dimensional Humanoid (rllab), are exceptionally difficult to solve with off-policy algorithms~\citep{duan2016benchmarking}. The stability of the algorithm also plays a large role in performance: easier tasks make it more practical to tune hyperparameters to achieve good results, while the already narrow basins of effective hyperparameters become prohibitively small for the more sensitive algorithms on the hardest benchmarks, leading to poor performance~\citep{gu2016q}. 

We compare our method to deep deterministic policy gradient (DDPG)~\citep{lillicrap2015continuous}, an algorithm that is regarded as one of the more efficient off-policy deep RL methods~\citep{duan2016benchmarking}; proximal policy optimization (PPO)~\citep{schulman2017proximal}, a stable and effective on-policy policy gradient algorithm; and soft Q-learning (SQL)~\citep{haarnoja2017reinforcement}, a recent off-policy algorithm for learning maximum entropy policies. Our SQL implementation also includes two Q-functions, which we found to improve its performance in most environments. 
We additionally compare to twin delayed deep deterministic policy gradient algorithm (TD3) \citep{fujimoto2018addressing}, using the author-provided implementation. This is an extension to DDPG, proposed concurrently to our method, that first applied the double Q-learning trick to continuous control along with other improvements. We have included trust region path consistency learning (Trust-PCL)~\citep{nachum2017trust} and two other variants of SAC in \autoref{app:benchmarks}. We turned off the exploration noise for evaluation for DDPG and PPO. For maximum entropy algorithms, which do not explicitly inject exploration noise, we either evaluated with the exploration noise (SQL) or use the mean action (SAC). The source code of our SAC implementation\footnote{\href{http://github.com/haarnoja/sac}{github.com/haarnoja/sac}} and videos\footnote{\href{sites.google.com/view/soft-actor-critic}{sites.google.com/view/soft-actor-critic}} are available online.

\subsection{Comparative Evaluation}
\label{sec:evaluation}
\autoref{fig:training_curves} shows the total average return of evaluation rollouts during training for DDPG, PPO, and TD3. We train five different instances of each algorithm with different random seeds, with each performing one evaluation rollout every 1000 environment steps. The solid curves corresponds to the mean and the shaded region to the minimum and maximum returns over the five trials.

The results show that, overall, SAC performs comparably to the baseline methods on the easier tasks and outperforms them on the  harder tasks with a large margin, both in terms of learning speed and the final performance. For example, DDPG fails to make any progress on Ant-v1, Humanoid-v1, and Humanoid (rllab), a result that is corroborated by prior work~\citep{gu2016q,duan2016benchmarking}. SAC also learns considerably faster than PPO as a consequence of the large batch sizes PPO needs to learn stably on more high-dimensional and complex tasks. 
Another maximum entropy RL algorithm, SQL, can also learn all tasks, but it is slower than SAC and has worse asymptotic performance. 
The quantitative results attained by SAC in our experiments also compare very favorably to results reported by other methods in prior work~\citep{duan2016benchmarking,gu2016q,henderson2017deep}, indicating that both the sample efficiency and final performance of SAC on these benchmark tasks exceeds the state of the art. All hyperparameters used in this experiment for SAC are listed in \aref{app:hypers}.

\subsection{Ablation Study}
\label{sec:ablations}

The results in the previous section suggest that algorithms based on the maximum entropy principle can outperform conventional RL methods on challenging tasks such as the humanoid tasks. In this section, we further examine which particular components of SAC are important for good performance. We also examine how sensitive SAC is to some of the most important hyperparameters, namely reward scaling and target value update smoothing constant.
\begin{figure}
\begin{centering}
    \includegraphics[width=0.75\columnwidth, trim={0 0 5mm 0mm}, clip]{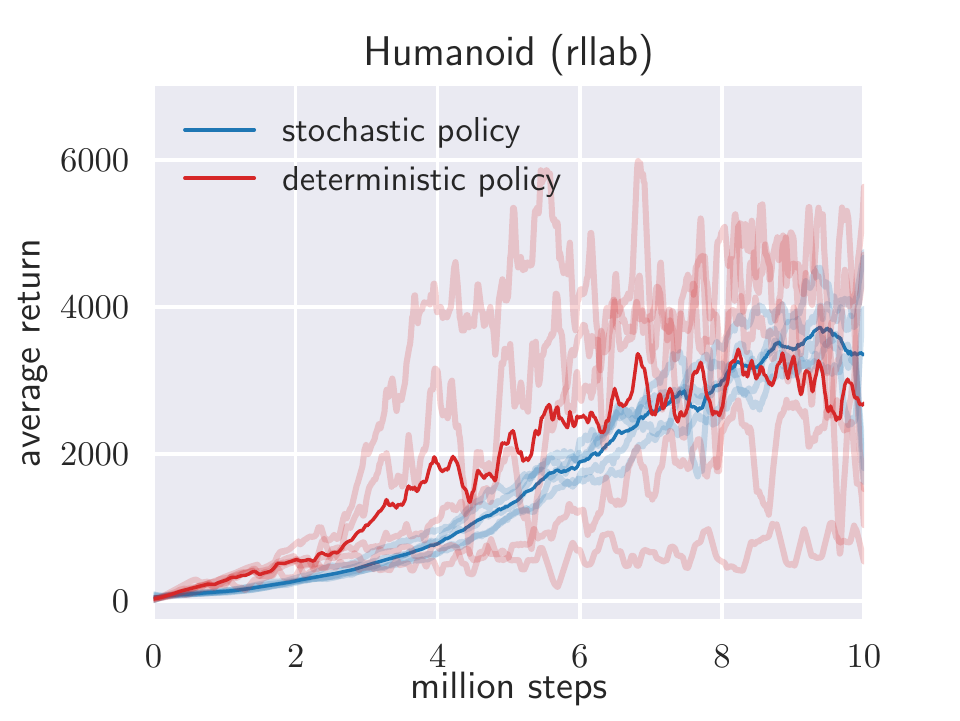}
    \caption{\small Comparison of SAC (blue) and a deterministic variant of SAC (red) in terms of the stability of individual random seeds on the Humanoid (rllab) benchmark. The comparison indicates that stochasticity can  stabilize training as the variability between the seeds becomes much higher with a deterministic policy.}
\label{fig:humanoid_seeds}
\end{centering}
\end{figure}

\vspace{-0.1in}
\paragraph{Stochastic vs. deterministic policy.} Soft actor-critic learns stochastic policies via a maximum entropy objective. The entropy appears in both the policy and value function. In the policy, it prevents premature convergence of the policy variance (\autoref{eq:policy_objective}). In the value function, it encourages exploration by increasing the value of regions of state space that lead to high-entropy behavior (\autoref{eq:v_cost}). To compare how the stochasticity of the policy and entropy maximization affects the performance, we compare to a deterministic variant of SAC that does not maximize the entropy and that closely resembles DDPG, with the exception of having two Q-functions, using hard target updates, not having a separate target actor, and using fixed rather than learned exploration noise. \autoref{fig:humanoid_seeds} compares five individual runs with both variants, initialized with different random seeds. Soft actor-critic performs much more consistently, while the deterministic variant exhibits very high variability across seeds, indicating substantially worse stability. As evident from the figure, learning a stochastic policy with entropy maximization can drastically stabilize training. This becomes especially important with harder tasks, where tuning hyperparameters is challenging. In this comparison, we updated the target value network weights with hard updates, by periodically overwriting the target network parameters to match the current value network (see \autoref{app:benchmarks} for a comparison of average performance on all benchmark tasks).

\begin{figure*}[t]
    \centering
    \subfigure[Evaluation]{
        \includegraphics[width=0.31\textwidth, trim={0 0 5mm 7.5mm}, clip]{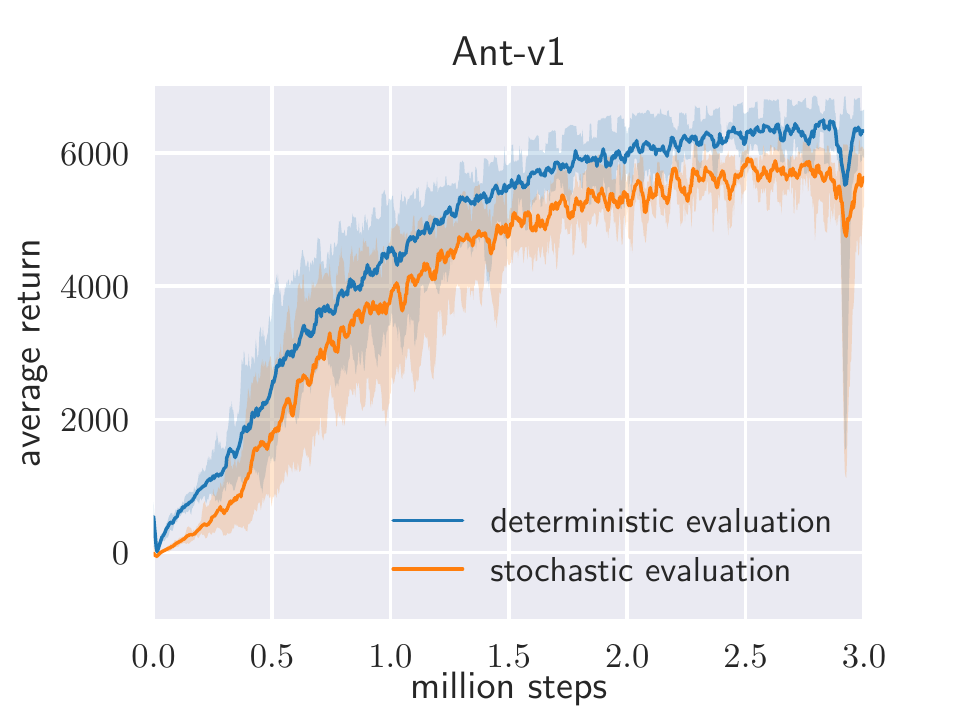}
        \label{fig:evaluation_ant}
    }
    \subfigure[Reward Scale]{
        \includegraphics[width=0.31\textwidth, trim={0 0 5mm 7.5mm}, clip]{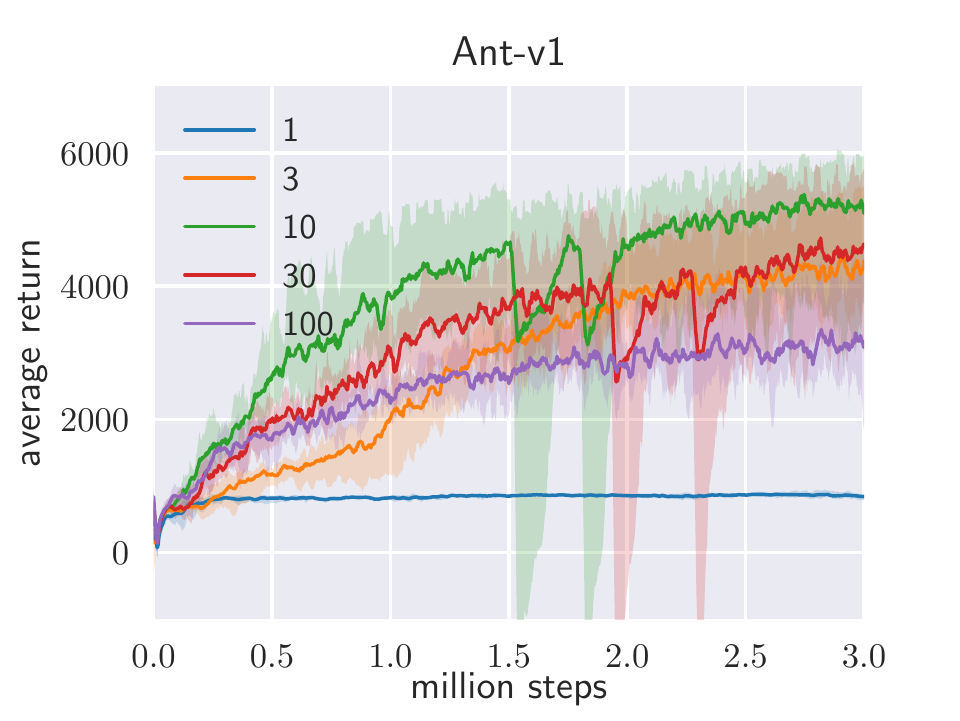}
        \label{fig:reward_scale_ant}
    }
    \subfigure[Target Smoothing Coefficient ($\tau$)]{
        \includegraphics[width=0.31\textwidth, trim={0 0 5mm 7.5mm}, clip]{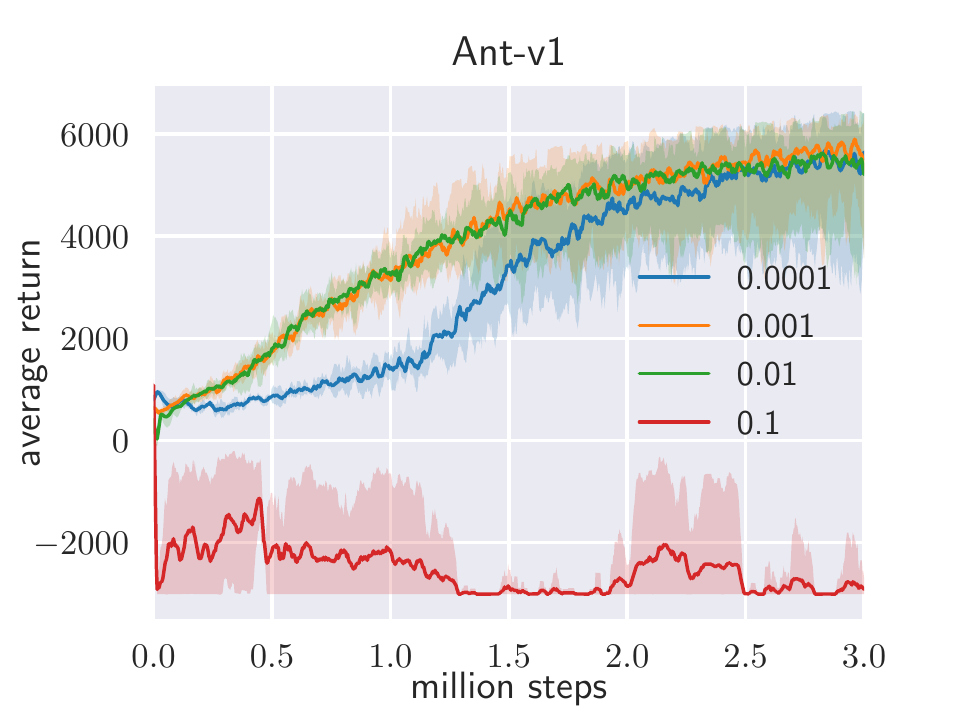}
        \label{fig:soft_target_ant}
    }
  	\caption{Sensitivity of soft actor-critic to selected hyperparameters on Ant-v1 task. (a) Evaluating the policy using the mean action generally results in a higher return. Note that the policy is trained to maximize also the entropy, and the mean action does not, in general, correspond the optimal action for the maximum return objective. (b) Soft actor-critic is sensitive to reward scaling since it is related to the temperature of the optimal policy. The optimal reward scale varies between environments, and should be tuned for each task separately. (c) Target value smoothing coefficient $\tau$ is used to stabilize training. Fast moving target (large $\tau$) can result in instabilities (red), whereas slow moving target (small $\tau$) makes training slower (blue).}
	\label{fig:sweeps}
\end{figure*}

\vspace{-0.1in}
\paragraph{Policy evaluation.} Since SAC converges to stochastic policies, it is often beneficial to make the final policy deterministic at the end for best performance. For evaluation, we approximate the maximum a posteriori action by choosing the mean of the policy distribution. \autoref{fig:evaluation_ant} compares training returns to evaluation returns obtained with this strategy indicating that deterministic evaluation can yield better performance. It should be noted that all of the training curves depict the sum of rewards, which is different from the objective optimized by SAC and other maximum entropy RL algorithms, including SQL and Trust-PCL, which maximize also the entropy of the policy.

\vspace{-0.1in}
\paragraph{Reward scale.} Soft actor-critic is particularly sensitive to the scaling of the reward signal, because it serves the role of the temperature of the energy-based optimal policy and thus controls its stochasticity. Larger reward magnitudes correspond to lower entries. \autoref{fig:reward_scale_ant} shows how learning performance changes when the reward scale is varied: For small reward magnitudes, the policy becomes nearly uniform, and consequently fails to exploit the reward signal, resulting in substantial degradation of performance. For large reward magnitudes, the model learns quickly at first, but the policy then becomes nearly deterministic, leading to poor local minima due to lack of adequate exploration. With the right reward scaling, the model balances exploration and exploitation, leading to faster learning and better asymptotic performance. In practice, we found reward scale to be the only hyperparameter that requires tuning, and its natural interpretation as the inverse of the temperature in the maximum entropy framework provides good intuition for how to adjust this parameter.

\vspace{-0.1in}
\paragraph{Target network update.} It is common to use a separate target value network that slowly tracks the actual value function to improve stability. We use an exponentially moving average, with a smoothing constant $\tau$, to update the target value network weights as common in the prior work~\cite{lillicrap2015continuous,mnih2015human}. A value of one corresponds to a hard update where the weights are copied directly at every iteration and zero to not updating the target at all. In \autoref{fig:soft_target_ant}, we compare the performance of SAC when $\tau$ varies. Large $\tau$ can lead to instabilities while small $\tau$ can make training slower. However, we found the range of suitable values of $\tau$ to be relatively wide and we used the same value (0.005) across all of the tasks. In \autoref{fig:training_curves_all} (\autoref{app:benchmarks}) we also compare to another variant of SAC, where instead of using exponentially moving average, we copy over the current network weights directly into the target network every 1000 gradient steps. We found this variant to benefit from taking more than one gradient step between the environment steps, which can improve performance but also increases the computational cost.

\newpage
\section{Conclusion}

We present soft actor-critic (SAC), an off-policy maximum entropy deep reinforcement learning algorithm that provides sample-efficient learning while retaining the benefits of entropy maximization and stability. Our theoretical results derive soft policy iteration, which we show to converge to the optimal policy. From this result, we can formulate a soft actor-critic algorithm, and we empirically show that it outperforms state-of-the-art model-free deep RL methods, including the off-policy DDPG algorithm and the on-policy PPO algorithm. In fact, the sample efficiency of this approach actually exceeds that of DDPG by a substantial margin. Our results suggest that stochastic, entropy maximizing reinforcement learning algorithms can provide a promising avenue for improved robustness and stability, and further exploration of maximum entropy methods, including methods that incorporate second order information (e.g., trust regions~\citep{schulman2015trust}) %
or more expressive policy classes is an exciting avenue for future work.

\section*{Acknowledgments}
We would like to thank Vitchyr Pong for insightful discussions and help in implementing our algorithm as well as providing the DDPG baseline code; Ofir Nachum for offering support in running Trust-PCL experiments; and George Tucker for his valuable feedback on an early version of this paper. This work was supported by Siemens and Berkeley DeepDrive.

\newpage

\FloatBarrier

\bibliography{refs}

\begin{thebibliography}{36}
\providecommand{\natexlab}[1]{#1}
\providecommand{\url}[1]{\texttt{#1}}
\expandafter\ifx\csname urlstyle\endcsname\relax
  \providecommand{\doi}[1]{doi: #1}\else
  \providecommand{\doi}{doi: \begingroup \urlstyle{rm}\Url}\fi

\bibitem[Barto et~al.(1983)Barto, Sutton, and Anderson]{barto1983neuronlike}
Barto, A.~G., Sutton, R.~S., and Anderson, C.~W.
\newblock Neuronlike adaptive elements that can solve difficult learning
  control problems.
\newblock \emph{IEEE transactions on systems, man, and cybernetics}, pp.\
  834--846, 1983.

\bibitem[Bhatnagar et~al.(2009)Bhatnagar, Precup, Silver, Sutton, Maei, and
  Szepesv{\'a}ri]{bhatnagar2009convergent}
Bhatnagar, S., Precup, D., Silver, D., Sutton, R.~S., Maei, H.~R., and
  Szepesv{\'a}ri, C.
\newblock Convergent temporal-difference learning with arbitrary smooth
  function approximation.
\newblock In \emph{Advances in Neural Information Processing Systems (NIPS)},
  pp.\  1204--1212, 2009.

\bibitem[Brockman et~al.(2016)Brockman, Cheung, Pettersson, Schneider,
  Schulman, Tang, and Zaremba]{brockman2016openai}
Brockman, G., Cheung, V., Pettersson, L., Schneider, J., Schulman, J., Tang,
  J., and Zaremba, W.
\newblock Open{AI} gym.
\newblock \emph{arXiv preprint arXiv:1606.01540}, 2016.

\bibitem[Duan et~al.(2016)Duan, Chen, Schulman, and
  Abbeel]{duan2016benchmarking}
Duan, Y., Chen, X.~Houthooft, R., Schulman, J., and Abbeel, P.
\newblock Benchmarking deep reinforcement learning for continuous control.
\newblock In \emph{International Conference on Machine Learning (ICML)}, 2016.

\bibitem[Fox et~al.(2016)Fox, Pakman, and Tishby]{fox2015taming}
Fox, R., Pakman, A., and Tishby, N.
\newblock Taming the noise in reinforcement learning via soft updates.
\newblock In \emph{Conference on Uncertainty in Artificial Intelligence (UAI)},
  2016.

\bibitem[Fujimoto et~al.(2018)Fujimoto, van Hoof, and
  Meger]{fujimoto2018addressing}
Fujimoto, S., van Hoof, H., and Meger, D.
\newblock Addressing function approximation error in actor-critic methods.
\newblock \emph{arXiv preprint arXiv:1802.09477}, 2018.

\bibitem[Gruslys et~al.(2017)Gruslys, Azar, Bellemare, and
  Munos]{gruslys2017reactor}
Gruslys, A., Azar, M.~G., Bellemare, M.~G., and Munos, R.
\newblock The reactor: A sample-efficient actor-critic architecture.
\newblock \emph{arXiv preprint arXiv:1704.04651}, 2017.

\bibitem[Gu et~al.(2016)Gu, Lillicrap, Ghahramani, Turner, and Levine]{gu2016q}
Gu, S., Lillicrap, T., Ghahramani, Z., Turner, R.~E., and Levine, S.
\newblock Q-prop: Sample-efficient policy gradient with an off-policy critic.
\newblock \emph{arXiv preprint arXiv:1611.02247}, 2016.

\bibitem[Haarnoja et~al.(2017)Haarnoja, Tang, Abbeel, and
  Levine]{haarnoja2017reinforcement}
Haarnoja, T., Tang, H., Abbeel, P., and Levine, S.
\newblock Reinforcement learning with deep energy-based policies.
\newblock In \emph{International Conference on Machine Learning (ICML)}, pp.\
  1352--1361, 2017.

\bibitem[Hasselt(2010)]{hasselt2010double}
Hasselt, H.~V.
\newblock Double {Q}-learning.
\newblock In \emph{Advances in Neural Information Processing Systems (NIPS)},
  pp.\  2613--2621, 2010.

\bibitem[Heess et~al.(2015)Heess, Wayne, Silver, Lillicrap, Erez, and
  Tassa]{heess2015learning}
Heess, N., Wayne, G., Silver, D., Lillicrap, T., Erez, T., and Tassa, Y.
\newblock Learning continuous control policies by stochastic value gradients.
\newblock In \emph{Advances in Neural Information Processing Systems (NIPS)},
  pp.\  2944--2952, 2015.

\bibitem[Henderson et~al.(2017)Henderson, Islam, Bachman, Pineau, Precup, and
  Meger]{henderson2017deep}
Henderson, P., Islam, R., Bachman, P., Pineau, J., Precup, D., and Meger, D.
\newblock Deep reinforcement learning that matters.
\newblock \emph{arXiv preprint arXiv:1709.06560}, 2017.

\bibitem[Kingma \& Ba(2015)Kingma and Ba]{kingma2014adam}
Kingma, D. and Ba, J.
\newblock Adam: A method for stochastic optimization.
\newblock In \emph{International Conference for Learning Presentations (ICLR)},
  2015.

\bibitem[Levine \& Koltun(2013)Levine and Koltun]{levine2013guided}
Levine, S. and Koltun, V.
\newblock Guided policy search.
\newblock In \emph{International Conference on Machine Learning (ICML)}, pp.\
  1--9, 2013.

\bibitem[Levine et~al.(2016)Levine, Finn, Darrell, and Abbeel]{levine2016end}
Levine, S., Finn, C., Darrell, T., and Abbeel, P.
\newblock End-to-end training of deep visuomotor policies.
\newblock \emph{Journal of Machine Learning Research}, 17\penalty0
  (39):\penalty0 1--40, 2016.

\bibitem[Lillicrap et~al.(2015)Lillicrap, Hunt, Pritzel, Heess, Erez, Tassa,
  Silver, and Wierstra]{lillicrap2015continuous}
Lillicrap, T.~P., Hunt, J.~J., Pritzel, A., Heess, N., Erez, T., Tassa, Y.,
  Silver, D., and Wierstra, D.
\newblock Continuous control with deep reinforcement learning.
\newblock \emph{arXiv preprint arXiv:1509.02971}, 2015.

\bibitem[Mnih et~al.(2013)Mnih, Kavukcuoglu, Silver, Graves, Antonoglou,
  Wierstra, and Riedmiller]{mnih2013playing}
Mnih, V., Kavukcuoglu, K., Silver, D., Graves, A., Antonoglou, I., Wierstra,
  D., and Riedmiller, M.
\newblock Playing atari with deep reinforcement learning.
\newblock \emph{arXiv preprint arXiv:1312.5602}, 2013.

\bibitem[Mnih et~al.(2015)Mnih, Kavukcuoglu, Silver, Rusu, Veness, Bellemare,
  Graves, Riedmiller, Fidjeland, Ostrovski, et~al.]{mnih2015human}
Mnih, V., Kavukcuoglu, K., Silver, D., Rusu, A.~A., Veness, J., Bellemare,
  M.~G., Graves, A., Riedmiller, M., Fidjeland, A.~K., Ostrovski, G., et~al.
\newblock Human-level control through deep reinforcement learning.
\newblock \emph{Nature}, 518\penalty0 (7540):\penalty0 529--533, 2015.

\bibitem[Mnih et~al.(2016)Mnih, Badia, Mirza, Graves, Lillicrap, Harley,
  Silver, and Kavukcuoglu]{mnih2016asynchronous}
Mnih, V., Badia, A.~P., Mirza, M., Graves, A., Lillicrap, T.~P., Harley, T.,
  Silver, D., and Kavukcuoglu, K.
\newblock Asynchronous methods for deep reinforcement learning.
\newblock In \emph{International Conference on Machine Learning (ICML)}, 2016.

\bibitem[Nachum et~al.(2017{\natexlab{a}})Nachum, Norouzi, Xu, and
  Schuurmans]{nachum2017bridging}
Nachum, O., Norouzi, M., Xu, K., and Schuurmans, D.
\newblock Bridging the gap between value and policy based reinforcement
  learning.
\newblock In \emph{Advances in Neural Information Processing Systems (NIPS)},
  pp.\  2772--2782, 2017{\natexlab{a}}.

\bibitem[Nachum et~al.(2017{\natexlab{b}})Nachum, Norouzi, Xu, and
  Schuurmans]{nachum2017trust}
Nachum, O., Norouzi, M., Xu, K., and Schuurmans, D.
\newblock Trust-{PCL}: An off-policy trust region method for continuous
  control.
\newblock \emph{arXiv preprint arXiv:1707.01891}, 2017{\natexlab{b}}.

\bibitem[O'Donoghue et~al.(2016)O'Donoghue, Munos, Kavukcuoglu, and
  Mnih]{o2016pgq}
O'Donoghue, B., Munos, R., Kavukcuoglu, K., and Mnih, V.
\newblock {PGQ}: Combining policy gradient and {Q}-learning.
\newblock \emph{arXiv preprint arXiv:1611.01626}, 2016.

\bibitem[Peters \& Schaal(2008)Peters and Schaal]{peters2008reinforcement}
Peters, J. and Schaal, S.
\newblock Reinforcement learning of motor skills with policy gradients.
\newblock \emph{Neural networks}, 21\penalty0 (4):\penalty0 682--697, 2008.

\bibitem[Rawlik et~al.(2012)Rawlik, Toussaint, and
  Vijayakumar]{rawlik2012stochastic}
Rawlik, K., Toussaint, M., and Vijayakumar, S.
\newblock On stochastic optimal control and reinforcement learning by
  approximate inference.
\newblock \emph{Robotics: Science and Systems (RSS)}, 2012.

\bibitem[Schulman et~al.(2015)Schulman, Levine, Abbeel, Jordan, and
  Moritz]{schulman2015trust}
Schulman, J., Levine, S., Abbeel, P., Jordan, M.~I., and Moritz, P.
\newblock Trust region policy optimization.
\newblock In \emph{International Conference on Machine Learning (ICML)}, pp.\
  1889--1897, 2015.

\bibitem[Schulman et~al.(2017{\natexlab{a}})Schulman, Abbeel, and
  Chen]{schulman2017equivalence}
Schulman, J., Abbeel, P., and Chen, X.
\newblock Equivalence between policy gradients and soft {Q}-learning.
\newblock \emph{arXiv preprint arXiv:1704.06440}, 2017{\natexlab{a}}.

\bibitem[Schulman et~al.(2017{\natexlab{b}})Schulman, Wolski, Dhariwal,
  Radford, and Klimov]{schulman2017proximal}
Schulman, J., Wolski, F., Dhariwal, P., Radford, A., and Klimov, O.
\newblock Proximal policy optimization algorithms.
\newblock \emph{arXiv preprint arXiv:1707.06347}, 2017{\natexlab{b}}.

\bibitem[Silver et~al.(2014)Silver, Lever, Heess, Degris, Wierstra, and
  Riedmiller]{silver2014deterministic}
Silver, D., Lever, G., Heess, N., Degris, T., Wierstra, D., and Riedmiller, M.
\newblock Deterministic policy gradient algorithms.
\newblock In \emph{International Conference on Machine Learning (ICML)}, 2014.

\bibitem[Silver et~al.(2016)Silver, Huang, Maddison, Guez, Sifre, van~den
  Driessche, Schrittwieser, Antonoglou, Panneershelvam, Lanctot, Dieleman,
  Grewe, Nham, Kalchbrenner, Sutskever, Lillicrap, Leach, Kavukcuoglu, Graepel,
  and Hassabis]{silver2016mastering}
Silver, D., Huang, A., Maddison, C.~J., Guez, A., Sifre, L., van~den Driessche,
  G., Schrittwieser, J., Antonoglou, I., Panneershelvam, V., Lanctot, M.,
  Dieleman, S., Grewe, D., Nham, J., Kalchbrenner, N., Sutskever, I.,
  Lillicrap, T., Leach, M., Kavukcuoglu, K., Graepel, T., and Hassabis, D.
\newblock Mastering the game of go with deep neural networks and tree search.
\newblock \emph{Nature}, 529\penalty0 (7587):\penalty0 484--489, Jan 2016.
\newblock ISSN 0028-0836.
\newblock Article.

\bibitem[Sutton \& Barto(1998)Sutton and Barto]{sutton1998reinforcement}
Sutton, R.~S. and Barto, A.~G.
\newblock \emph{Reinforcement learning: An introduction}, volume~1.
\newblock MIT press Cambridge, 1998.

\bibitem[Thomas(2014)]{thomas2014bias}
Thomas, P.
\newblock Bias in natural actor-critic algorithms.
\newblock In \emph{International Conference on Machine Learning (ICML)}, pp.\
  441--448, 2014.

\bibitem[Todorov(2008)]{todorov2008general}
Todorov, E.
\newblock General duality between optimal control and estimation.
\newblock In \emph{IEEE Conference on Decision and Control (CDC)}, pp.\
  4286--4292. IEEE, 2008.

\bibitem[Toussaint(2009)]{toussaint2009robot}
Toussaint, M.
\newblock Robot trajectory optimization using approximate inference.
\newblock In \emph{International Conference on Machine Learning (ICML)}, pp.\
  1049--1056. ACM, 2009.

\bibitem[Williams(1992)]{williams1992simple}
Williams, R.~J.
\newblock Simple statistical gradient-following algorithms for connectionist
  reinforcement learning.
\newblock \emph{Machine learning}, 8\penalty0 (3-4):\penalty0 229--256, 1992.

\bibitem[Ziebart(2010)]{ziebart2010modeling}
Ziebart, B.~D.
\newblock \emph{Modeling purposeful adaptive behavior with the principle of
  maximum causal entropy}.
\newblock Carnegie Mellon University, 2010.

\bibitem[Ziebart et~al.(2008)Ziebart, Maas, Bagnell, and
  Dey]{ziebart2008maximum}
Ziebart, B.~D., Maas, A.~L., Bagnell, J.~A., and Dey, A.~K.
\newblock Maximum entropy inverse reinforcement learning.
\newblock In \emph{AAAI Conference on Artificial Intelligence (AAAI)}, pp.\
  1433--1438, 2008.

\end{thebibliography}
\bibliographystyle{icml2018}

\newpage
\onecolumn
\appendix

\section{Maximum Entropy Objective}
\label{app:objective}
The exact definition of the discounted maximum entropy objective is complicated by the fact that, when using a discount factor for policy gradient methods, we typically do not discount the state distribution, only the rewards. In that sense, discounted policy gradients typically do not optimize the true discounted objective. Instead, they optimize average reward, with the discount serving to reduce variance, as discussed by \citet{thomas2014bias}. However, we can define the objective that \emph{is} optimized under a discount factor as
\begin{align}
J(\policy) = \sum_{t=0}^\infty \E{(\st,\at) \sim \rho_\policy}{ \sum_{l=t}^\infty \discount^{l-t} \E{\state_l\sim\pdyn,\action_l\sim\policy}{ \reward(\st,\at) + \alpha \ent(\policy(\voidarg|\st))|\st,\at}}.
\end{align}
This objective corresponds to maximizing the discounted expected reward and entropy for future states originating from every state-action tuple $(\st,\at)$ weighted by its probability $\rho_\pi$ under the current policy.

\section{Proofs}

\subsection{\autoref{lem:soft_policy_evaluation}}
\label{app:lem_soft_policy_evaluation}

\textbf{\autoref{lem:soft_policy_evaluation}} (Soft Policy Evaluation).\textit{
Consider the soft Bellman backup operator $\mathcal{T}^\policy$ in \autoref{eq:soft_bellman_backup_op} and a mapping $Q^0: \sspace \times \aspace\rightarrow \reals$ with $|\aspace|<\infty$, and define $\Q^{k+1} = \mathcal{T}^\policy \Q^k$. Then the sequence $Q^k$ will converge to the soft Q-value of $\policy$ as $k\rightarrow \infty$.}
\begin{proof}
Define the entropy augmented reward as $\reward_\policy(\st, \at) \triangleq \reward(\st, \at)  + \E{\stp\sim\pdyn}{\entropy\left(\policy(\voidarg|\stp)\right)}$ and rewrite the update rule as 
\begin{align}
\Q(\st, \at) \leftarrow \reward_\policy(\st, \at) + \discount \E{\stp\sim\pdyn,\atp \sim \policy}{Q(\stp, \atp)}
\end{align}
and apply the standard convergence results for policy evaluation~\citep{sutton1998reinforcement}. The assumption $|\aspace|<\infty$ is required to guarantee that the entropy augmented reward is bounded.
\end{proof}

\subsection{\autoref{lem:policy_improvement}}
\label{app:lem_policy_improvement}

\textbf{\autoref{lem:policy_improvement}} (Soft Policy Improvement).\textit{
Let $\policy_\mathrm{old} \in \Pi$ and let $\policy_\mathrm{new}$ be the optimizer of the minimization problem defined in \autoref{eq:constrainted_policy_fitting}. Then $\Q^{\policy_\mathrm{new}}(\st, \at) \geq \Q^{\policy_\mathrm{old}}(\st, \at)$ for all $(\st, \at) \in \sspace\times\aspace$ with $|\aspace|<\infty$.}
\begin{proof}
Let $\policyold\in \Pi$ and let $\Q^\policyold$ and $\V^\policyold$ be the corresponding soft state-action value and soft state value, and let $\policynew$ be defined as 
\begin{align}
\policynew(\voidarg|\st) &= \arg \min_{\policy' \in \Pi}\kl{\policy'(\voidarg|\st)}{\exp\left(Q^\policyold(\st,\voidarg) - \log Z^\policyold(\st)\right)}\notag\\
 &= \arg\min_{\policy'\in\Pi}J_\policyold(\policy'(\voidarg|\st)).
\end{align}
It must be the case that $J_\policyold(\policynew(\voidarg|\st)) \leq J_\policyold(\policyold(\voidarg|\st))$, since we can always choose $\policynew = \policyold\in\Pi$. Hence
\begin{align}
\resizebox{1\textwidth}{!}{$
\E{\at\sim\policynew}{\log \policynew(\at|\st) - Q^\policyold(\st, \at) + \log Z^\policyold(\st)}
\leq \E{\at\sim\policyold}{\log \policyold(\at|\st) - Q^\policyold(\st,\at) + \log Z^\policyold(\st)}$},
\end{align}
and since partition function $Z^\policyold$ depends only on the state, the inequality reduces to
\begin{align}
\E{\at\sim\policynew}{Q^\policyold(\st, \at) - \log \policynew(\at|\st)} \geq V^\policyold(\st).
\label{eq:soft_value_bound}
\end{align}
Next, consider the soft Bellman equation:
\begin{align}
Q^\policyold(\st, \at) &= \reward(\st, \at) + \discount\E{\stp\sim\pdyn}{V^\policyold(\stp)}\notag\\
&\leq \reward(\st, \at) + \discount\E{\stp\sim\pdyn}{\E{\atp\sim\policynew}{Q^\policyold(\stp, \atp) - \log \policynew(\atp|\stp)}}\notag\\
&\ \  \vdots\notag\\
& \leq Q^\policynew(\st, \at),
\end{align}
where we have repeatedly expanded $Q^\policyold$ on the RHS by applying the soft Bellman equation and the bound in \autoref{eq:soft_value_bound}. Convergence to $\Q^\policynew$ follows from \autoref{lem:soft_policy_evaluation}.
\end{proof}

\subsection{\autoref{the:soft_policy_iteration}}
\label{app:the_soft_policy_iteration}

\textbf{\autoref{the:soft_policy_iteration}} (Soft Policy Iteration). \textit{
Repeated application of soft policy evaluation and soft policy improvement to any $\policy\in\Pi$ converges to a policy $\policy\opt$ such that $Q^{\policy\opt}(\st, \at) \geq Q^{\policy}(\st, \at)$ for all $\policy \in \Pi$ and $(\st, \at) \in \sspace\times\aspace$, assuming $|\aspace|<\infty$.}
\begin{proof}
Let $\policy_i$ be the policy at iteration $i$. By \autoref{lem:policy_improvement}, the sequence $Q^{\policy_i}$ is monotonically increasing. Since $Q^\policy$ is bounded above for $\policy \in \Pi$ (both the reward and entropy are bounded), the sequence converges to some $\policy\opt$. We will still need to show that $\policy\opt$ is indeed optimal. At convergence, it must be case that $J_\policyopt(\policyopt(\voidarg|\st)) < J_\policyopt(\policy(\voidarg|\st))$ for all $\policy\in\Pi$, $\policy\neq \policy\opt$. Using the same iterative argument as in the proof of \autoref{lem:policy_improvement}, we get $Q^\policyopt(\st, \at) > Q^\policy(\st, \at)$ for all $(\st, \at)\in \sspace\times\aspace$, that is, the soft value of any other policy in $\Pi$ is lower than that of the converged policy. Hence $\policyopt$ is optimal in $\Pi$.
\end{proof}

\section{Enforcing Action Bounds}
\label{app:action_bounds}
We use an unbounded Gaussian as the action distribution. However, in practice, the actions needs to be bounded to a finite interval. To that end, we
apply an invertible squashing function 
($\tanh$) to the Gaussian samples, and employ the change of variables formula to compute the likelihoods of the bounded actions. In the other words, let $\urv\in\reals^D$ be a random variable and $\mu(\urv|\state)$ the corresponding density with infinite support. Then $\action = \tanh(\urv)$, where $\tanh$ is applied elementwise, is a random variable with support in $(-1, 1)$ with a density given by
\begin{align}
\policy (\action|\state) &= \mu(\urv|\state)\left|\det \left(\frac{\mathrm{d}\action}{\mathrm{d}\urv} \right)\right|^{-1}.
\end{align}
Since the Jacobian $\nicefrac{\mathrm{d}\action}{\mathrm{d}\urv} = \mathrm{diag}(1 - \tanh^2(\urv))$ is diagonal, the log-likelihood has a simple form 
\begin{align}
\log\policy (\action|\state) &= \log \mu(\urv|\state) - \sum_{i=1}^D\log\left(1 - \tanh^2(u_i)\right),
\end{align}
where $u_i$ is the $i^\mathrm{th}$ element of $\urv$.

\newpage
\section{Hyperparameters}
\label{app:hypers}
\autoref{tab:shared_params} lists the common SAC parameters used in the comparative evaluation in \autoref{fig:training_curves} and \autoref{fig:training_curves_all}. \autoref{tab:env_params} lists the reward scale parameter that was tuned for each environment.

\begin{table}[H]
\renewcommand{\arraystretch}{1.1}
\centering
\caption{SAC Hyperparameters}
\label{tab:shared_params}
\vspace{1mm}
\begin{tabular}{l l| l }
\toprule
\multicolumn{2}{l|}{Parameter} &  Value\\
\midrule
\multicolumn{2}{l|}{\it{Shared}}& \\
& optimizer &Adam \citep{kingma2014adam}\\
& learning rate & $3 \cdot 10^{-4}$\\
& discount ($\discount$) &  0.99\\
& replay buffer size & $10^6$\\
& number of hidden layers (all networks) & 2\\
& number of hidden units per layer & 256\\
& number of samples per minibatch & 256\\
& nonlinearity & ReLU\\
\midrule
\multicolumn{2}{l|}{\it{SAC}}& \\
& target smoothing coefficient ($\tau$)& 0.005\\
& target update interval & 1\\
& gradient steps & 1\\
\midrule
\multicolumn{2}{l|}{\it{SAC (hard target update)}}& \\
& target smoothing coefficient ($\tau$)& 1\\
& target update interval & 1000\\
& gradient steps (except humanoids)& 4\\
& gradient steps (humanoids)& 1\\
\bottomrule
\end{tabular}
\end{table}

\begin{table}[H]
\renewcommand{\arraystretch}{1.1}
\centering
\caption{SAC Environment Specific Parameters}
\label{tab:env_params}
\vspace{1mm}
\begin{tabular}{ l l l l l }
\toprule
Environment 		&Action Dimensions	&Reward Scale\\
\midrule
Hopper-v1 			&3		& 5\\
Walker2d-v1 		&6 		& 5\\
HalfCheetah-v1 		&6		& 5\\
Ant-v1 				&8		& 5\\
Humanoid-v1 	    &17 	& 20\\
Humanoid (rllab) 	&21 	& 10\\
\bottomrule
\end{tabular}
\end{table}

\newpage
\section{Additional Baseline Results}
\label{app:benchmarks}
\begin{figure*}[h]
    \centering
    \subfigure[Hopper-v1]{
        \includegraphics[width=0.31\textwidth, trim={0 0 5mm 7.5mm}, clip]{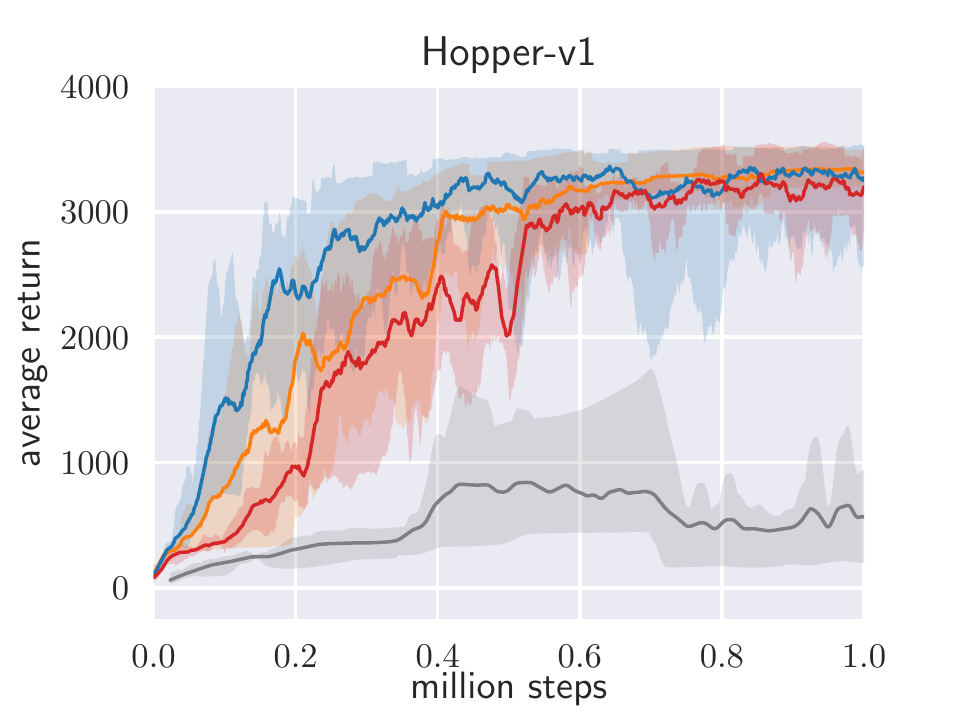}
    }
  	\subfigure[Walker2d-v1]{
        \includegraphics[width=0.31\textwidth, trim={0 0 5mm 7.5mm}, clip]{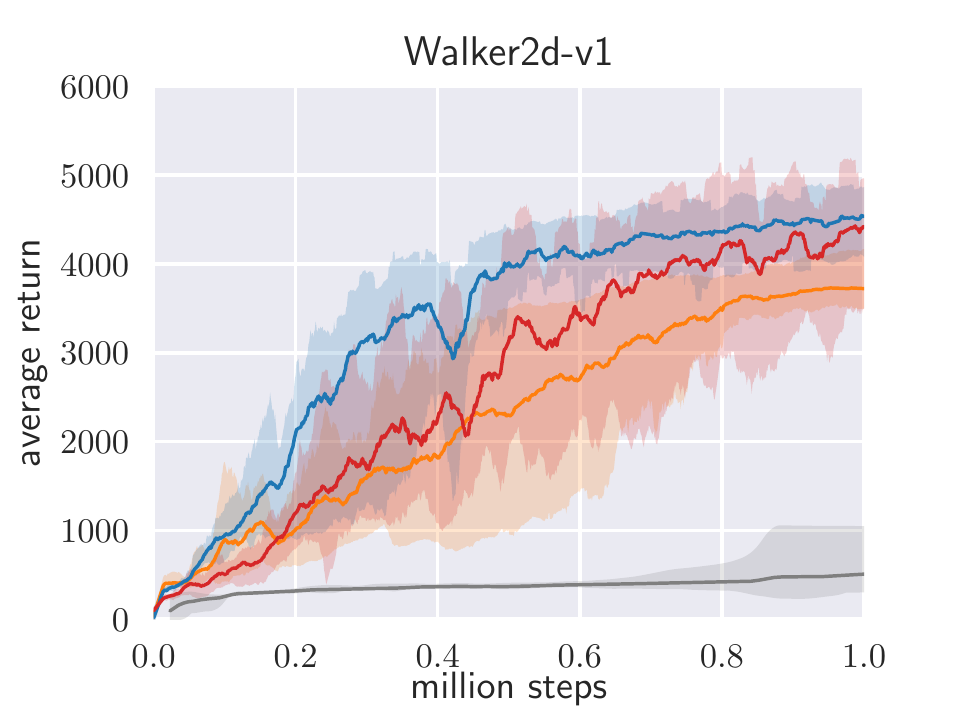}
    }
    \subfigure[HalfCheetah-v1]{
        \includegraphics[width=0.31\textwidth, trim={0 0 5mm 7.5mm}, clip]{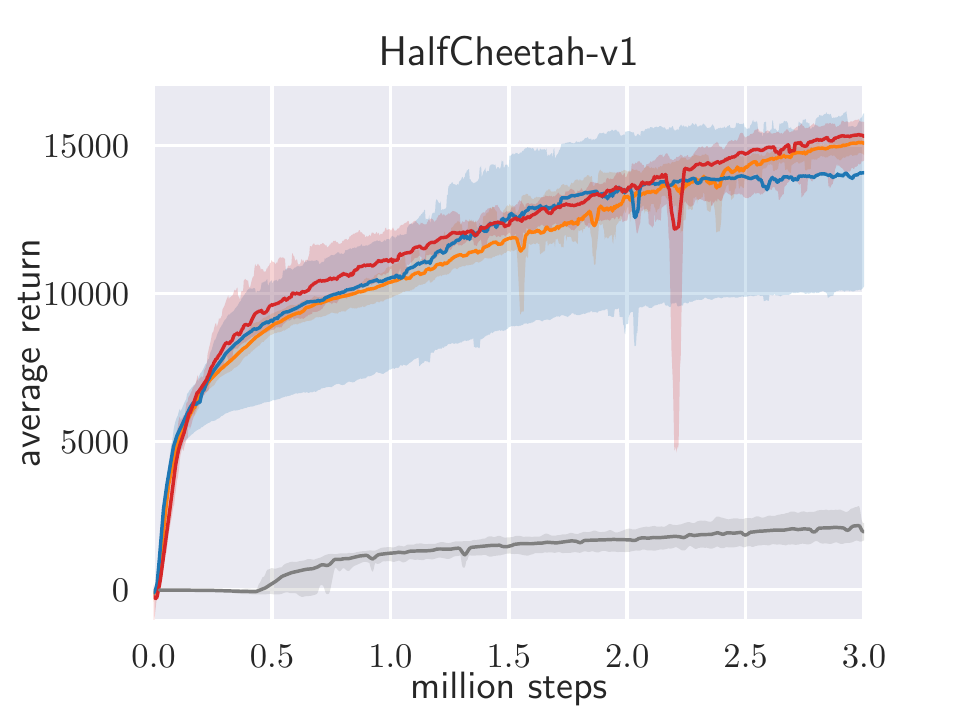}
    }
    \subfigure[Ant-v1]{
        \includegraphics[width=0.31\textwidth, trim={0 0 5mm 7.5mm}, clip]{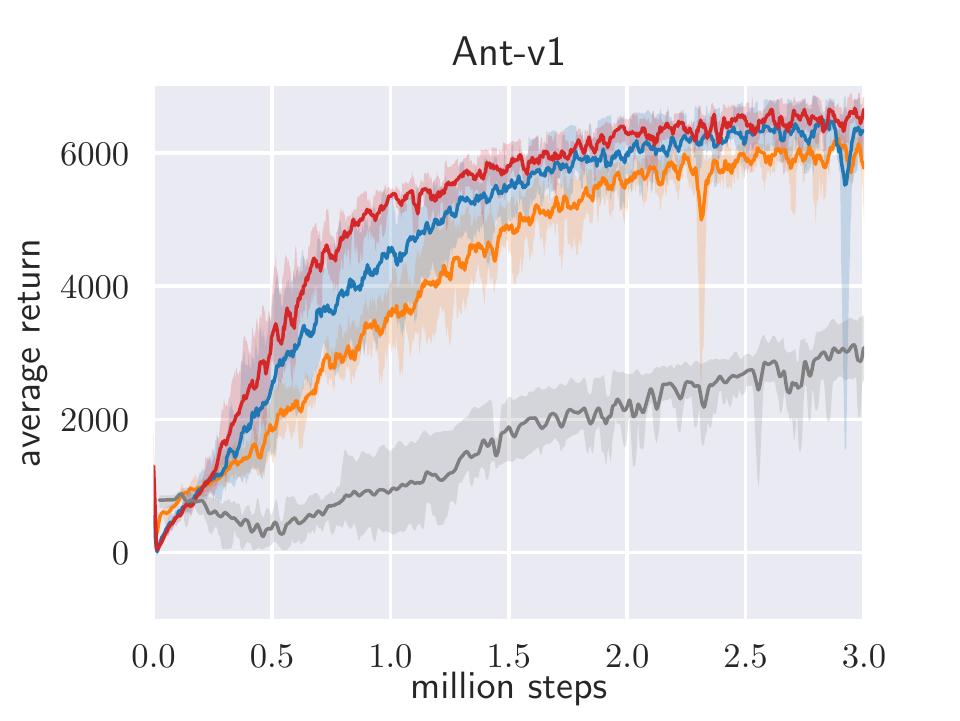}
    }
    \subfigure[Humanoid-v1]{
        \includegraphics[width=0.31\textwidth, trim={0 0 5mm 7.5mm}, clip]{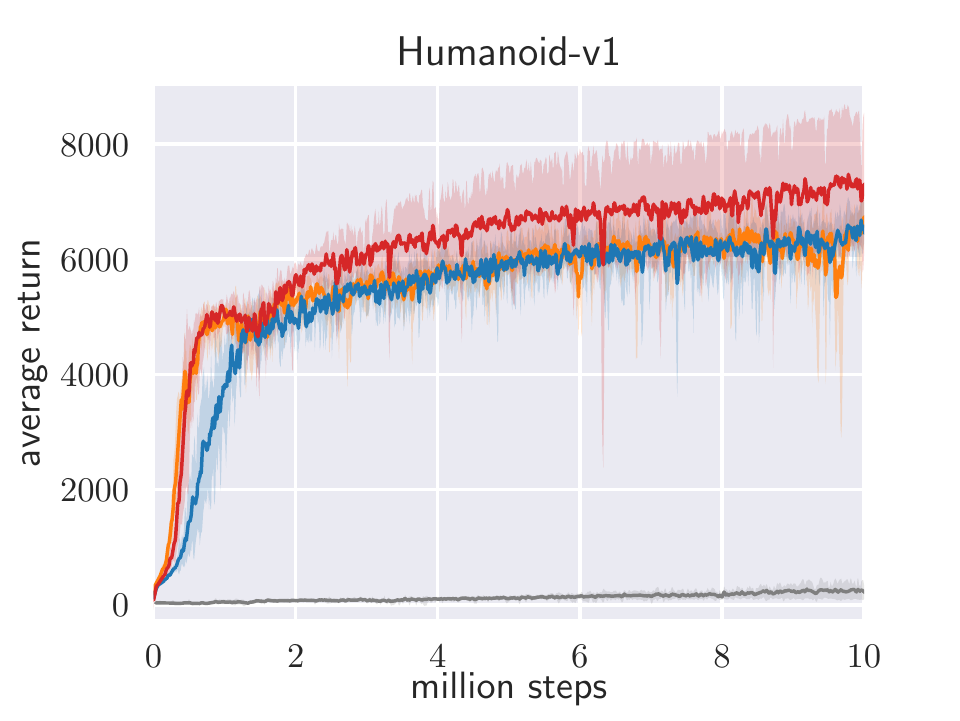}
    }
    \subfigure[Humanoid (rllab)]{
        \includegraphics[width=0.31\textwidth, trim={0 0 5mm 7.5mm}, clip]{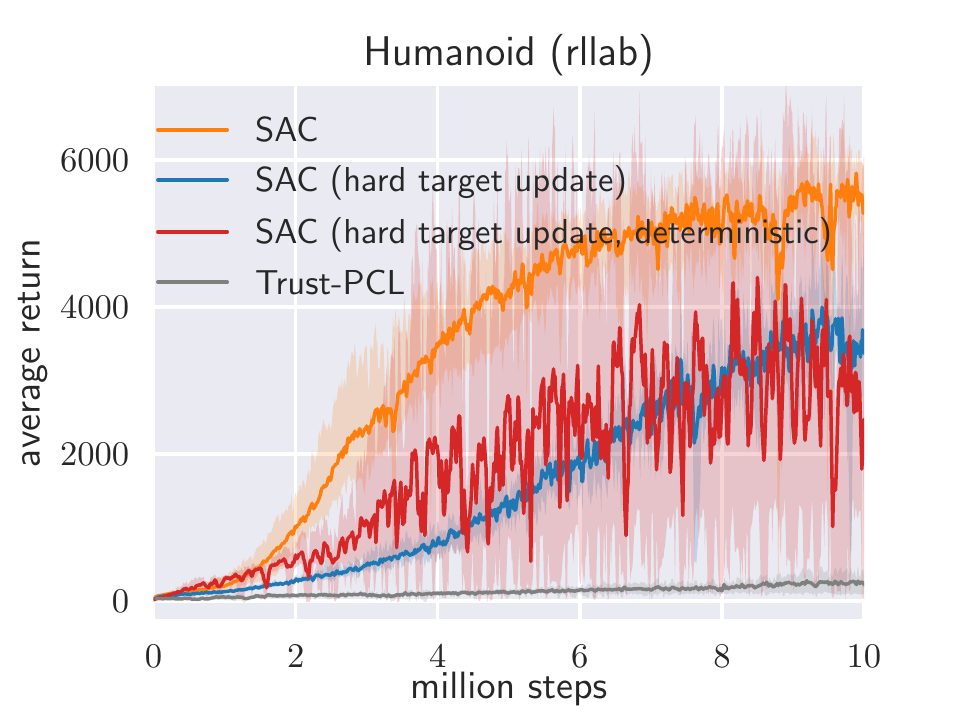}
    }
    \caption{\small Training curves for additional baseline (Trust-PCL) and for two SAC variants. Soft actor-critic with hard target update (blue) differs from standard SAC in that it copies the value function network weights directly every 1000 iterations, instead of using exponentially smoothed average of the weights. The deterministic ablation (red) uses a deterministic policy with fixed Gaussian exploration noise, does not use a value function, drops the entropy terms in the actor and critic function updates, and uses hard target updates for the target Q-functions. It is equivalent to DDPG that uses two Q-functions, hard target updates, and removes the target actor.}
	\label{fig:training_curves_all}
 \end{figure*}

\autoref{fig:training_curves_all} compares SAC to Trust-PCL (\autoref{fig:training_curves_all}. Trust-PC fails to solve most of the task within the given number of environment steps, although it can eventually solve the easier tasks \citep{nachum2017trust} if ran longer. The figure also includes two variants of SAC: a variant that periodically copies the target value network weights directly instead of using exponentially moving average, and a deterministic ablation which assumes a deterministic policy in the value update (\autoref{eq:v_gradient}) and the policy update (\autoref{eq:policy_gradient}), and thus strongly resembles DDPG with the exception of having two Q-functions, using hard target updates, not having a separate target actor, and using  fixed exploration noise rather than learned. Both of these methods can learn all of the tasks and they perform comparably to SAC on all but Humanoid (rllab) task, on which SAC is the fastest. 

\end{document}